\documentclass[lettersize,journal]{IEEEtran}
\usepackage[numbers,sort&compress]{natbib}
\usepackage{blindtext}
\usepackage[bottom]{footmisc}
\usepackage{graphicx}
\usepackage{epstopdf}
\usepackage{subcaption}
\usepackage{amsmath,amsfonts,bm}
\usepackage{amssymb}
\usepackage{amsthm}
\usepackage{cases}
\usepackage{setspace}
\usepackage{multirow}
\usepackage{array}
\usepackage[caption=false,font=normalsize,labelfont=sf,textfont=sf]{subfig}
\usepackage{textcomp} 
\usepackage{stfloats}
\usepackage{url}
\usepackage{verbatim}
\def\BibTeX{{\rm B\kern-.05em{\sc i\kern-.025em b}\kern-.08em
    T\kern-.1667em\lower.7ex\hbox{E}\kern-.125emX}}
\usepackage{balance}
\usepackage{mathtools}
\usepackage[dvipsnames]{xcolor}
\usepackage[ruled, linesnumbered]{algorithm2e}
\SetKwComment{Comment}{// }{}
\usepackage{tabulary}
\usepackage{booktabs}
\usepackage{tabularx} 
\usepackage{svg}

\allowdisplaybreaks

\usepackage{adjustbox}

\begin{document}

\title{Efficient Training of Large-Scale AI Models through Federated Mixture-of-Experts: A System-Level Approach}

\author{
            {Xiaobing Chen,
            Boyang Zhang,
		Xiangwei Zhou,
            Mingxuan Sun,
            Shuai Zhang,
            Songyang Zhang, 
            and Geoffrey Ye Li ~\IEEEmembership{Fellow,~IEEE}
            }
    \thanks{X. Chen and X. Zhou are with the Division of Electrical and Computer Engineering, Louisiana State University, Baton Rouge, LA 70803, USA (e-mail: \{xchen87, xwzhou\}@lsu.edu).}    
    \thanks{B. Zhang and M. Sun are with the Division of Computer Science and Engineering, Louisiana State University, Baton Rouge, LA 70803, USA (e-mail: \{bzhang29, msun11\}@lsu.edu).}
    \thanks{Shuai Zhang is with the Department of Data Science, New Jersey Institute of Technology, Newark, NJ 07102, USA (e-mail: sz457@njit.edu).}
    \thanks{Songyang Zhang is with the Department of Electrical and Computer Engineering, University of Louisiana at Lafayette, Lafayette, LA 70504, USA (e-mail: songyang.zhang@louisiana.edu).}
    \thanks{G. Y. Li is with the Department of Electrical and Electronic Engineering, Imperial College London, London SW7 2AZ, UK (e-mail: Geoffrey.Li@Imperial.ac.uk).}
}



\maketitle

\begin{abstract}
The integration of Federated Learning (FL) and Mixture-of-Experts (MoE) presents a compelling pathway for training more powerful, large-scale artificial intelligence models (LAMs) on decentralized data while preserving privacy. However, efficient federated training of these complex MoE-structured LAMs is hindered by significant system-level challenges, particularly in managing the interplay between heterogeneous client resources and the sophisticated coordination required for numerous specialized experts. This article highlights a critical, yet underexplored concept: the absence of robust quantitative strategies for dynamic client-expert alignment that holistically considers varying client capacities and the imperative for system-wise load balancing. Specifically, we propose a conceptual system design for intelligent client-expert alignment that incorporates dynamic fitness scoring, global expert load monitoring, and client capacity profiling. By tackling these systemic issues, we can unlock more scalable, efficient, and robust training mechanisms {with fewer communication rounds for convergence}, paving the way for the widespread deployment of large-scale federated MoE-structured LAMs in edge computing {with ultra-high communication efficiency}.
\end{abstract}


\section{Introduction}
\IEEEPARstart{T}{he} relentless advancement of large-scale artificial intelligence models (LAMs) has triggered numerous emerging technologies, such as large language models and vision foundation models, which offer unprecedented capabilities across a multitude of applications from scientific analysis to practical problem solving, including biomedical image processing, remote sensing and wireless communications \cite{chang2024survey}. 
Despite offering superior performance and transformative capabilities, 
these LAMs, with billions of parameters, demand immense computational resources and memory for both training and inference, limiting their deployment in practical scenarios, particularly in edge computing systems with a constrained computational capacity. Furthermore, the vast data for training LAMs are often geometrically distributed across various organizations or user devices, resulting in growing privacy concerns and logistical challenges related to data aggregation, which do not lend themselves to conventional centralized learning.
Although distributed learning provides potential solutions, significant communication and computation overheads are introduced, particularly with streaming data and heterogeneous hardware capabilities \cite{liu2024split}.

To enhance the efficiency and privacy preservation for training LAMs in distributed systems, two powerful paradigms have recently gained prominence: Federated Learning (FL) and Mixture-of-Experts (MoE) \cite{shazeer2017outrageously}. By enabling collaborative model training without sharing raw data, FL offers a compelling path to address data privacy and learning efficiency in distributed edge computing \cite{mcmahan2017communication}. On the other hand, MoE architectures provide an effective strategy to scale LAMs to enormous sizes while maintaining computational tractability by dividing the LAM into numerous smaller, specialized ``expert" sub-networks, as shown in Fig. \ref{fig:application_scenario}, where only a subset of experts is activated during inference. Through sparsely activating optimal experts, MoE-structured LAM can significantly reduce the computation cost and enhance task performance. Thus, integration of FL and MoE can be a natural solution to enable efficient and effective LAMs in edge computing while preserving privacy. However, as an emerging technology, the federated training of MoE-structured LAM remains an ill-posed and underexplored problem. In this work, we will introduce the key concepts of FL for MoE-structured LAM and propose a conceptual system design.
\subsection{Scope and Overview}
\begin{figure*}[t]
    \includegraphics[width=\linewidth]{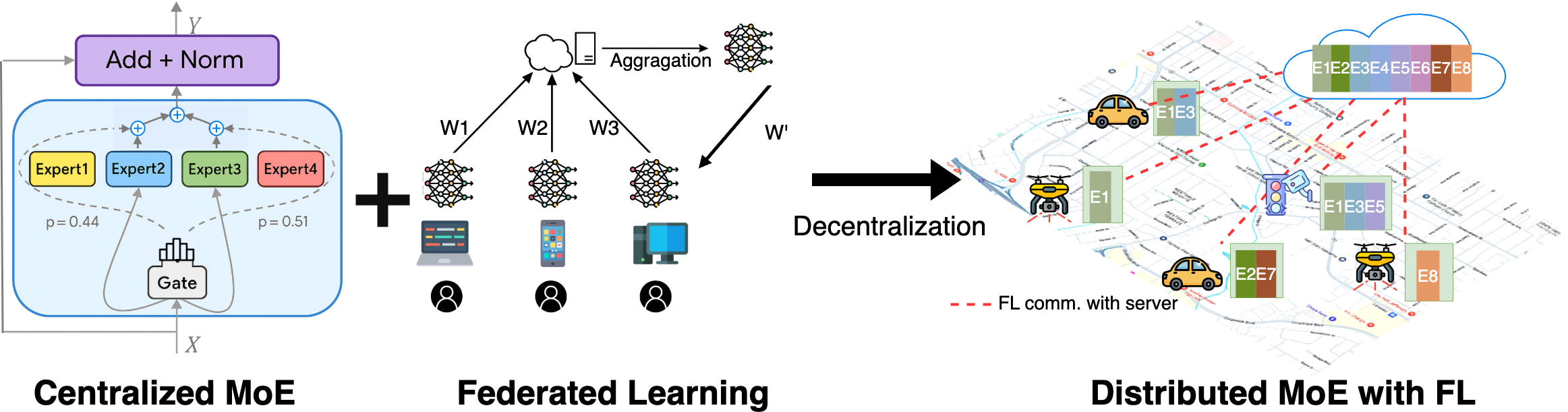}
    \caption{Proposed system of FL-enabled, MoE-structured LAM for edge computing: 1) Centralized MoE consists of a gating network and several expert networks, where the gating function routes each data sample to its most suitable expert for processing; 2) FL aims to collaboratively train the objective neural network by sharing gradients/models instead of raw data, which is capable of preserving data privacy; 3) Example of distributed MoE with FL:
    In a smart city sensor system, different edge devices such as vehicles, drones, and intelligent traffic infrastructure act as clients. Each client can be tasked with training a specific subset of experts (e.g., E1, E3, E5 for the traffic sensor in the figure) from a large, comprehensive MoE model based on its unique local data. Through FL, these clients collaboratively refine a global MoE LAM capable of complex tasks, such as intelligent traffic management or public safety monitoring.}
    \label{fig:application_scenario}
\end{figure*}

This work focuses on the federated training of MoE-structured LAM for edge computing. As shown in Fig.\ref{fig:application_scenario}, the distributed MoE-structured LAM system consists of a central server for global management and several clients (e.g., edge devices or deployed sensors) with different computational capacities and corresponding local data. Through FL, these clients aim to collaboratively refine a global MoE-structured LAM without sharing raw data. Due to the limitation of computational resources, each client can only be assigned to train a specific subset of experts simultaneously. After federated training, the most suitable experts are assigned to each client during inference. This synergistic design allows for efficient training and effective deployment of more powerful and specialized MoE-structured LAMs for real-world problems across diverse domains while preserving data privacy.

Despite the intuition of this distributed MoE-structured LAM system design, realizing its efficient and scalable training in a federated manner remains an underexplored problem. Resulting from the demand for specialization of each expert and heterogeneous data distribution, the first challenge lies in the personalized training of experts for the appropriate clients. Moreover, the varying occupancy of limited local computational resources across different clients, together with the dynamic communication channel condition, further increases the difficulty of client-expert alignment, motivating the development of resource-aware strategies for MoE training. Although some existing studies have discussed the deployment and gating of MoE-structured LAM in edge computing, considering channel quality \cite{song2025mixture} or end-to-end latency \cite{xue2024wdmoe}, they only focus on training in a centralized setup, while the practicalities of federated training for MoE remain challenging.

\subsection{Contents and Structures}

This article aims to delineate the critical area of FL-enabled MoE-structured LAMs for edge computing. Specifically, we highlight the current lack of quantitative and holistic analysis and optimization strategies to address the intricate interplay among three key system-level factors: 1) the heterogeneous system capacities of clients; 2) the dynamic client-expert alignment to match appropriate clients to each expert; and 3) the overarching system-wise load balancing for efficient and effective MoE training. To address the aforementioned challenges, we propose a conceptual system design for dynamic client-expert alignment considering both varying client capacities and system-wise load balancing within the FL setup. Experimental results validate the promising potential for the proposed system design.

The remainder of this work is organized as follows. We first review the foundations of FL and MoE, together with the current stage of integrating FL and MoE in Section \ref{sec:prelim}. Then, we illustrate the major challenges of FL-enabled MoE from the systematic perspective and propose our conceptual design in Section \ref{sec:system}. Following the discussion of additional challenges and future opportunities in Section \ref{sec:challenge}, we summarize our work in Section \ref{sec:conclusion}.

\section{Federated Learning and Mixture-of-Experts}\label{sec:prelim}
Before stepping into the FL-enabled, MoE-structured LAM design for edge computing, we first provide an overview of the current stage of FL and MoE, together with their integration.
\subsection{Federated Learning}
FL is a decentralized learning paradigm in which several clients collaboratively train a shared LAM without exchanging their raw local data, coordinated by a central server \cite{mcmahan2017communication}. In a standard FL process, clients compute the gradients/model updates based on their local datasets, after which these updates are aggregated by the server to refine the global model. This iterative process allows the global model to learn from a diverse range of data sources while protecting data privacy.
The primary allure of FL for LAM training lies in its inherent ability to preserve the confidentiality of sensitive information by keeping data on client devices without direct sharing. Beyond privacy, FL can harness the power of diverse data sources and reduce the burdens associated with huge raw data transmission, storage, and management.

Despite its successes, the practical deployment of FL for LAMs faces substantial challenges rooted in client heterogeneity. This heterogeneity manifests in two primary forms: system heterogeneity and statistical heterogeneity \cite{10589575}. System heterogeneity refers to the wide disparities in computational resources, network connectivity, and power availability for each client. Statistical heterogeneity arises from the non-IID nature of data across clients, reflected by the distribution and composition of data, e.g., class distributions and feature skew, 
which can vary dramatically across clients and potentially hinder model convergence and performance.

Another bottleneck for FL of LAM lies in the communication overhead, which inherently involves a large number of parameters. The frequent transmission of model updates between potentially millions of clients and the central server can lead to significant latency and consume substantial network bandwidth. This communication bottleneck is especially critical in resource-constrained environments, such as wireless networks or clients constrained by limited data plans, severely impacting the scalability and practicality of federated training for large models.

\subsection{Mixture-of-Experts}

As shown in Fig. \ref{fig:application_scenario}, MoE employs multiple specialized ``expert" sub-networks to achieve high capacity and computational efficiency, allowing for vast parameter counts without a proportional increase in computation \cite{shazeer2017outrageously}. Unlike the conventional single monolithic dense model, MoE contains a ``gating network" (or router), which dynamically selects a small subset (e.g., top-k) of experts per input, enabling sparse activation of a fraction of total parameters per step.
This sparsely-gating mechanism allows for the efficient and scalable implementation of LAMs while keeping the computational cost per example manageable and nearly constant, regardless of the total number of experts. Furthermore, the division of labor among experts naturally fosters specialization, where different experts learn to handle different types of data or subtasks, leading to improved model accuracy and interpretability.

In traditional centralized training setups, MoE-structured LAMs present unique systemic challenges of expert assignment. A primary concern is ensuring a balanced computational load across all experts to prevent inefficient training and suboptimal performance, where some experts are over-utilized while others are under-utilized. An effective routing strategy shall be capable of selecting the most relevant experts and generalizing well to unseen data, which remains a critical challenge from the learning perspective. These intrinsic challenges of MoE architectures are further amplified in the context of FL by introducing ``client", where load balancing appears to play an important role in all aspects of client, expert, and client-expert.

\begin{table*}[t]
    \centering 
    \caption{Summary of existing federated MoE methods. Client-expert alignment refers to whether the server dynamically assigns experts to clients. Client capacity denotes whether client system heterogeneity is considered.}
    \label{tab:fedmoe_summary}
    \begin{adjustbox}{width=\textwidth, totalheight=\textheight, keepaspectratio}
    \begin{tabularx}{\textwidth}{
        >{\raggedright\arraybackslash}p{0.18\textwidth}| 
        >{\raggedright\arraybackslash}p{0.18\textwidth} 
        >{\raggedright\arraybackslash}p{0.22\textwidth} 
        >{\raggedright\arraybackslash}p{0.16\textwidth} 
        >{\raggedright\arraybackslash}p{0.15\textwidth} 
    }
    \toprule
    \textbf{Method} &
    \textbf{Personalized or Global Focus?} &
    \textbf{Client-Expert Alignment?} &
    \textbf{Client Capacity?} &
    \textbf{System-Wise Load Balancing?} \\
    \midrule

    \textbf{FedJETs} \cite{dun2023fedjets} &
    Personalized &
    Based on routers &
    No &
    None\\

    \textbf{FedMix} \cite{reisser2021federated} &
    Hybrid &
    None &
    No &
    None \\

    \textbf{FedMoE-DA} \cite{zhan2024fedmoe} &
    Hybrid &
    P2P expert aggregation &
    Partial&
    None \\

    \textbf{FedMoE} \cite{mei2024fedmoe} &
    Hybrid &
    Heuristic search &
    Partial &
    None \\

    \textbf{pFedMoAP} \cite{luo2024mixture}&
    Personalized &
    None &
    No &
    None \\

    \textbf{pFedMoE} \cite{yi2024pfedmoe}&
    Personalized &
    None &
    No &
    None \\
    \bottomrule
    \end{tabularx}
    \end{adjustbox}
\end{table*}

\subsection{Current Stage of Federated MoE}
Recognizing the complementary strengths of FL and MoE, recent interest has been attracted by their integration, aiming to develop sophisticated models that benefit from both distributed data training and expert-based specialization.
Recently, a variety of approaches have emerged, including FedJETs \cite{dun2023fedjets}, FedMix \cite{reisser2021federated}, FedMoE-DA \cite{zhan2024fedmoe}, FedMoE \cite{mei2024fedmoe}, pFedMoAP \cite{luo2024mixture}, and pFedMoE \cite{yi2024pfedmoe}, each proposing unique ways to harness MoE structures to address prevalent challenges such as data heterogeneity and the need for personalization.
Existing federated MoE methods are summarized from the server-side optimization perspective in Table \ref{tab:fedmoe_summary}.

Most existing FL-MoE frameworks focus on personalized federated learning (PFL), adapting MoE principles to provide clients with customized models or experts. Systems, such as pFedMoE \cite{yi2024pfedmoe} and pFedMoAP \cite{luo2024mixture}, enable clients to blend global knowledge with local MoE adaptations, while FedJETs \cite{dun2023fedjets} route client data to specialized experts for tailored model behavior. Another category of frameworks aims to collaboratively train multiple distinct experts that are managed or integrated by a central server to form a more unified, server-side multi-expert model. For example, FedMix \cite{reisser2021federated} trains an ensemble of specialized models with clients choosing relevant members. FedMoE \cite{mei2024fedmoe} aims to construct an optimal sub-MoE for each client and enable the server to provide global expert recommendations. Other FL-MoE structures, such as FedMoE-DA \cite{zhan2024fedmoe}, also allow for client-specific expert training with peer-to-peer (P2P) sharing, indicating progress towards more comprehensive server-centric multiexpert systems in FL.
These pioneering systems have effectively demonstrated the utility of FL-MoE in managing statistical heterogeneity and enabling model adaptation. 

Despite the aforementioned advantages, the predominant focus of the existing integration of FL and MoE lies in demonstrating feasibility, achieving personalization, and managing data diversity, while an insightful quantitative analysis of efficiently training large-scale, server-centric MoE-structured LAMs in distributed edge computing remains underexplored. Although some system-level concerns, such as client selection, resource constraints, and client-side workload management, are touched upon (e.g., in FedJETs \cite{dun2023fedjets}, FedMix \cite{reisser2021federated}, FedMoE \cite{mei2024fedmoe}, FedMoE-DA \cite{zhan2024fedmoe}), these considerations often appear as secondary outcomes. 
The ignorance of constraints in available resources and system-level load balancing emerges as a critical challenge in deploying an FL-enabled MoE system in practical edge computing.

\section{Enabling FL of MoE-Structured LAM: A system-level approach}\label{sec:system}
We now discuss the key aspects to enable efficient federated training of MoE-structured LAM in edge computing, together with an exemplary system design.

\subsection{Key Factors from System Perspective}
The central thesis of this work is to improve the federated training of MoE-structured LAM from the system perspective, which focuses on addressing the lack of quantitative strategies for dynamic client-expert alignment, holistically considering the heterogeneous client capacities and system-wise expert load balancing. In particular, we shall focus on three system-level factors as follows.

\subsubsection{Client System Heterogeneity}
Most existing FL-MoE frameworks contribute to addressing the heterogeneity of data statistics from the learning perspective while ignoring the client system heterogeneity from the system perspective. In realistic FL of edge computing systems,
client devices usually exhibit vast differences in computational power, available memory, and network bandwidth. These varying capacities directly constrain their ability to participate effectively in training specific experts or subsets of experts within a large server-side MoE-structured LAM. For instance, memory-limited clients cannot load large amounts of experts, even though they may contain high-quality data. Additionally, low-bandwidth clients struggle with multi-round gradient transmission to ensure a timely update. How to quantitatively model these capacity impacts and system dynamics emerges as a critical challenge in enabling efficient FL of MoE-structured LAM.

\subsubsection{Client-Expert Alignment}
Impacted by client system heterogeneity, client-expert alignment is another critical concern in the federated MoE-structured LAM systems.
Although the gating mechanisms in conventional MoEs favor sending data to the most relevant experts for specialization, an optimal alignment strategy between experts and clients should be jointly optimized, considering the relevance of the data for the specialization of experts, the individual computational and communication capabilities of the clients, and the overall training needs and current state of the global MoE-structured LAM. This necessitates the future exploration of developing adaptive mechanisms to dynamically adjust client-expert assignments based on fluctuating client availability, resource profiles, and the learning progress of different experts.

\subsubsection{System-Wise Load Balancing}
System-wise load balancing across all experts is another critical, but often overlooked, hurdle intertwined with client-expert alignment in the federated MoE-structured LAMs. In centralized MOE training, load balancing ensures that all experts receive sufficient training samples. However, in FL, this effect is compounded by client heterogeneity and partial participation. For example, some experts might disproportionately receive more updates from frequently available clients, while others, particularly those specialized in rarer data types or tasks, become undertrained. This imbalance can severely degrade the performance and generalization capabilities of the MoE-structured LAM. Thus, sophisticated mechanisms are required to ensure that each expert receives sufficient, diverse, and timely training contributions from a dynamically fluctuating pool of clients, considering their varied capacities and data characteristics.

\subsubsection{Summary}
Ultimately, the crux of this unaddressed need lies in the absence of frameworks that can quantitatively model the complex trade-offs and enable the co-optimization of these interconnected systemic factors, including the heterogeneous client capacity, dynamic client-expert alignment, and system-wise expert load balance. An efficient federated training scheme is expected to support the adoption of a holistic system-level design in federated MoE-structured LAM to avoid drawbacks, including inefficient resource utilization, slow convergence, biased expert training, or the inability to effectively train the full scale of the intended MoE model.

\subsection{Proposed System Design}

\begin{figure}
    \includegraphics[width=\linewidth]{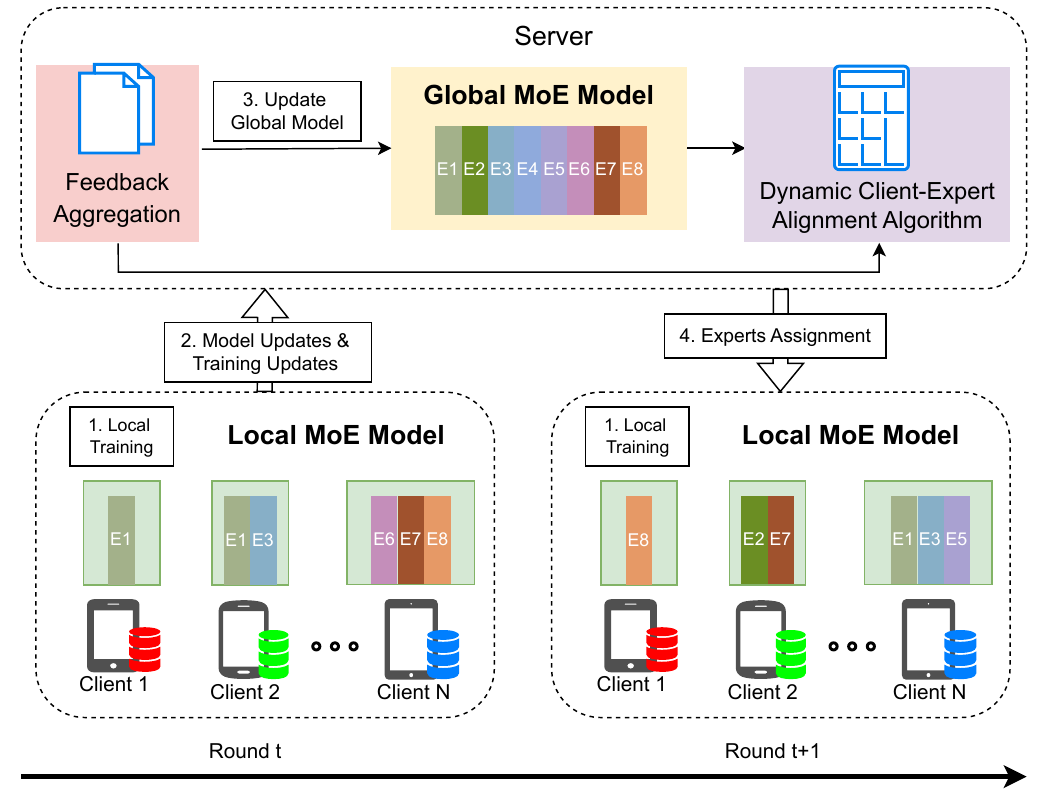}
    \captionsetup{justification=justified, singlelinecheck=false}
    \caption{Architecture of federated MoE-structured LAMs with client-expert alignment: 1) Local client initializes its model and uploads it to the server; 2) The server aggregates the updates and determines the client-expert alignment in the next round; 3) Upon receiving the new expert assignment, the client updates the models and continues steps above.}
    \label{fig:diagram_fedmoe}
\end{figure}

To address the aforementioned challenges of efficiently training MoE-structured LAMs in federated setups, we propose an exemplary system design centered on efficient client-expert alignment, as shown in Fig. \ref{fig:diagram_fedmoe}. This system, featuring continuously updated metrics and profiling mechanisms, considers all three key factors, e.g., heterogeneous client capacities, the evolving fitness of experts for specific client data, and the necessity of balanced training load across all experts, whose details
are illustrated as follows.

\subsubsection{Quantifying Client-Expert Fitness}
At the heart of our alignment strategy is the ability to dynamically assess how well an expert's specialization matches a client's local data characteristics. To achieve this, the system maintains a \textbf{Client-Expert Fitness Score} for each client-expert pair. This score estimates the suitability or quality of a particular expert for a specific client. A higher score indicates a better fit, implying that the expert is likely to perform well or be more frequently selected by the client's internal routing mechanism for its specific data.
A dynamic \textbf{Client-Expert Fitness Score} quantifies an expert's suitability for a client's data. This score can be designed based on client feedback during post-training, where a ``reward" (e.g., based on low error, frequent client-side expert selection) updates the score via an Exponential Moving Average (EMA) to reflect recent performance while retaining historical context. Non-interaction may gradually decrease the score.
\begin{figure*}[t]
    \includegraphics[width=\linewidth]{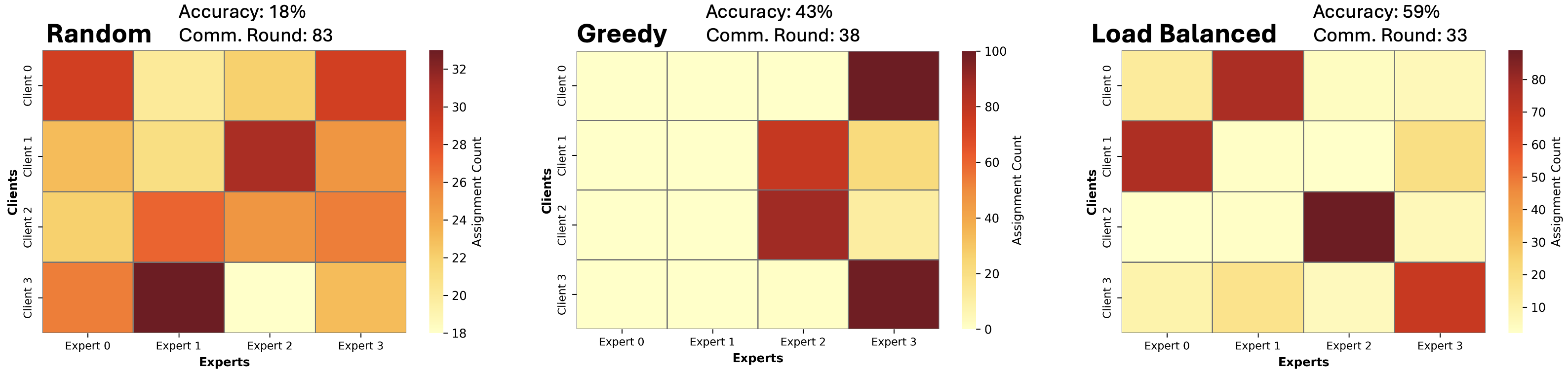}
    \captionsetup{justification=justified, singlelinecheck=false}
    \caption{Conceptual comparison of three client-expert assignment strategies under a non-IID data setting in CIFAR10 (darker color refers to more assignments). (a) \textbf{Random} ($Accuracy=18\%, Communication\_Round=83$): Inconsistent assignments lead to poor model specialization. (b) \textbf{Greedy} ($Accuracy=43\%, Communication\_Round=38$): A single popular expert is overloaded, creating an unbalanced model. (c) \textbf{Load-Balanced} ($Accuracy=59\%, Communication\_Round=33$): Clients are aligned with suitable experts, resulting in a well-trained and high-performing model. }
    \label{fig:compare_algo}
\end{figure*}
\subsubsection{Monitoring Global Expert Training Load}
To ensure that all experts within the global MoE model receive adequate and balanced training contributions, preventing scenarios where some experts are over-trained while others are neglected, the system maintains an \textbf{Expert Usage Score}, which tracks the cumulative training load or activity level for each expert across the entire federation of clients.

Similar to the fitness score, this expert usage score is updated after each round using an averaging technique that gives more weight to recent activity. It incorporates the total usage of an expert by all clients that trained it in that round (for example, by summing up the number of data samples or computational effort contributed by clients to that expert). A decay factor in the averaging process controls how quickly past usage information becomes less relevant, effectively defining a time window over which an expert's training load is considered most important for balancing purposes. A high usage score for an expert would indicate that it is currently receiving substantial training attention system-wide.

\subsubsection{Client Capacity Profiling}
A cornerstone of effective resource-aware alignment is understanding the capabilities of each participating client. The system incorporates \textbf{Client Capacity Profiling}, which maintains a profile for each client. This profile quantifies key aspects of the system resources for each client, detailed as follows.
\begin{itemize}
    \item Computational Capacity: This refers to the estimated processing power of each client, such as CPU or GPU speed and the number of available processing cores. This factor influences how quickly a client can perform local training on one or more experts.
    \item Memory Availability: This is the available RAM on the client device, which dictates the number of experts that a client can simultaneously load into memory and train.
    \item Network Conditions: The network condition considers the estimated bandwidth and network latency, which affect the time taken for a client to receive expert models from the server and transmit its updates back.
\end{itemize}

The capacity information can be self-reported, gathered by monitoring agents, or estimated by the server from historical performance (e.g., task completion times). This profile helps set practical limits for the maximum number of experts trained in each client per round.

\subsubsection{Dynamic Client-Expert Alignment Algorithm}
The core of the proposed system is a \textbf{Dynamic Client-Expert Alignment Algorithm}. This algorithm intelligently assigns a subset of experts for each selected client to be trained in each communication round by leveraging the continuously updated client-expert fitness scores, the global expert usage scores, and the individual client capacity profiles.

To illustrate the importance of this algorithm, we take a federated scenario with non-IID data as an example, where the data on each client are uniquely suited to a specific expert. 
As depicted in Fig. \ref{fig:compare_algo}, different assignment strategies yield vastly different outcomes, and the load-balanced strategy can significantly reduce the communication overhead for LAM convergence to achieve superior performance, leading to higher bandwidth efficiency.
This highlights that an effective method shall assign clients to experts with high affinity (a high fitness score) while prioritizing under-trained experts (a low usage score) to achieve system-wise load balance.

The client-expert algorithm in our proposed federated MoE-structured LAM system is described as follows.
\begin{itemize}

\item First, the server identifies a set of candidate experts that the client could potentially train. This selection considers the capacity profile of each client to ensure, for example, sufficient memory for expert training.

\item Next, for each of these candidate experts, the algorithm calculates a composite assignment desirability score. This score is a composite score that integrates multiple metrics. For example, it would be positively influenced by a high client-expert fitness score and negatively influenced by a high global expert usage score. Weighting factors can be used to adjust the relative importance of client-expert fitness versus the need for system-wise load balancing.

\item Finally, based on these composite desirability scores, the algorithm performs a capacity-constrained expert assignment, which selects the top-ranked experts for the client to train, up to the maximum number of experts that the client can handle in that round, as determined by its capacity profile and other system-level constraints.
\end{itemize}

Through explicitly factoring in the global expert usage scores when making assignment decisions, this dynamic alignment actively works towards achieving system-wise load balancing, which naturally discourages the over-assignment of popular or easily trainable experts and encourages the distribution of training load more evenly across the entire pool of experts in the MoE-structured LAM. 

\section{Future Directions}\label{sec:challenge}
This section outlines some promising opportunities and future directions in the federated MoE-structured LAMs.
\subsection{Enabling Advanced Distributed Applications}
The successful development of federated MoE-structured LAM systems
can enable the collaborative training of massive specialized models on sensitive and decentralized data, accelerating research in fields, such as genomics or drug discovery, and paving the way for enhanced personalized services delivered at the edge, such as highly adaptive recommendation systems or sophisticated beam prediction for mobile users in high-mobility environments \cite{10892257}.
Furthermore, robust and scalable federated MoE-structured LAMs can significantly bolster the intelligence of large-scale industrial Internet-of-Things (IoT) systems by 
optimizing communication and computation resource management.
Considering the specific requirements of different applications, one promising future direction is to customize the client-expert alignment considering the special data distribution and domain knowledge. Another potential direction is to employ LAM with hardware to further enhance the implementation efficiency for specific applications.

\subsection{Computation Efficiency}
A primary challenge in practical LAM is the efficient management of the computational load on resource-constrained clients. To this end, lightweight federated MoE-structured LAM shall be explored. One potential solution is to design a more compact model, such as splitting a large global model into smaller sub-networks distributed to clients based on their hardware capacities \cite{liu2024split}. This method can be further combined with adaptive pruning, which dynamically removes unnecessary parameters to reduce computation and communication costs. Using importance metrics to determine layer-wise pruning levels can be another alternative to create sparse and efficient models tailored for each client.

\subsection{Communication Resource Efficiency}
In edge computing, particularly with wireless environments, network conditions directly impact the performance of a distributed MoE system, as the dynamic nature of channels introduces uncertainty and can degrade model outcomes \cite{xue2024wdmoe}. This creates an opportunity to design intelligent channel-aware gating mechanisms that improve expert selection. Such policies can make routing decisions by considering not only data-expert alignment but also real-time factors, including channel state information and end-to-end latency \cite{song2025mixture}. This approach prevents scenarios where a well-specialized expert is rendered ineffective by a poor communication link.

\subsection{Privacy, Fairness, and Trust}
Although FL can improve privacy preservation, the complexity of MoE systems introduces new dimensions to consider \cite{xue2024wdmoe}. Distributing model components may enhance security, as no single location holds the complete model. However, patterns of expert usage and assignment may leak client information, such as data similarity and membership leakage. Furthermore, fairness is tightly coupled with the challenge of load balancing. Without equitable scheduling, clients with limited resources could be neglected, leading to under-trained experts and biased outcomes. Developing strategies that ensure fair participation and prevent such biases is another crucial direction for future work.

\section{Conclusion}\label{sec:conclusion}
This article introduces the key concepts of novel federated MoE-structured LAMs for edge computing, together with an exemplary system-level design to enhance the {learning} effectiveness and {communication} efficiency of federated training. Specifically, we consider three key factors in the federated MoE: 1) client system heterogeneity; 2) dynamic client-expert alignment; and 3) system-wise load balancing for experts. Fundamentally, the incorporation of resource-aware client-expert assignment facilitates the federated MoE with better scalability and efficiency, achieving the tradeoff between data statistics and client resources. The advancements of federated MoE-structured LAM can lead to extensive applications in edge computing and AIoT. With the development of artificial intelligence, together with its hardware deployment, federated MoEs are expected to significantly benefit the next-generation mobile computing and communications.

\section{Acknowledgement}
This material is based upon works supported in part by the National Science Foundation under Grant No. 1943486, 2246757, 2315612, 2332011, and 2349878, and in part by a grant from BoRSF under contract LEQSF(2024-
27)-RD-B-03.

\bibliographystyle{IEEEtran}
\bibliography{references}

\section*{Biography}
\noindent \textbf{Xiaobing Chen} 
is currently pursuing the Ph.D. degree in the Division of Electrical and Computer Engineering, Louisiana State University, where he joined as a Graduate Research Assistant in 2021. His research interests include federated learning, privacy in machine learning, and optimization theory.

\vspace{2mm}
\noindent \textbf{Boyang Zhang} 
is currently pursuing the Ph.D. degree in the Division of Computer Science and Engineering, Louisiana State University, where he joined as a Graduate Research Assistant in 2022. His research interests include fairness in machine learning and federated learning.

\vspace{2mm}
\noindent \textbf{Xiangwei Zhou} received 
the Ph.D. degree in electrical and computer engineering from Georgia Institute of Technology, Atlanta, GA, USA, in 
2011. He 
is an Associate Professor with the Division of Electrical and Computer Engineering, Louisiana State University, Baton Rouge, LA, USA. 
He was the recipient of the Best Paper Award at the 2014 International Conference on Wireless Communications and Signal Processing and served as an Editor for the IEEE Transactions on Wireless Communications from 2013 to 2018.

\vspace{2mm}
\noindent \textbf{Mingxuan Sun} received 
the Ph.D. degree in computer science from Georgia Institute of Technology, Atlanta, GA, USA in 2012. She is an Associate Professor with the Division of Computer Science and Engineering, Louisiana State University, Baton Rouge, LA, USA. Her research interests include machine learning, information retrieval, and data mining. She is also interested in machine learning and AI applications in social informatics, security, and wireless communications. 

\vspace{2mm}
\noindent \textbf{Shuai Zhang} received 
the Ph.D. degree from Rensselaer Polytechnic Institute, Troy, NY, USA, in 2021, where he was a Postdoctoral Research Associate from 2022 to 2023. He is currently an Assistant Professor with the Department of Data Science, New Jersey Institute of Technology, Newark, NJ, USA. 
His research focuses on theoretical foundations of deep learning and development of principled, and efficient algorithms 
for AI applications.

\vspace{2mm}
\noindent \textbf{Songyang Zhang} received the Ph.D. degree in Department of Electrical and Computer Engineering from the University of California at Davis, Davis, CA, USA, in 2021, where he was a Postdoctoral Research Associate from August 2021 to July 2023. He is currently an Assistant Professor with the Department of Electrical and Computer Engineering in University of Louisiana at Lafayette, Lafayette, LA, USA. His current research interests include machine learning, signal processing, IoT intelligence and wireless communications. 

\vspace{2mm}
\noindent \textbf{Geoffrey Ye Li} is a Chair Professor at Imperial College London. He made fundamental contributions to orthogonal frequency division multiplexing (OFDM) for wireless communications, established a framework on resource cooperation in wireless networks, and introduced deep learning to communications. He was elected to Fellow of Royal Academy of Engineering (FREng) and IEEE Fellow and won 2024 IEEE Eric E. Sumner Award and 2019 IEEE ComSoc Edwin Howard Arm-strong Achievement Award.
\end{document}